\documentclass{article}

\usepackage{graphicx}
\usepackage{subfigure} 

\usepackage{natbib}
\usepackage[noend]{algorithmic}
\usepackage{algorithmic}
\usepackage{hyperref}

\usepackage[accepted]{icml2012}

\usepackage[latin2]{inputenc}
\usepackage[T1]{fontenc}
\icmltitlerunning{Automated Word Puzzle Generation via Topic Dictionaries}

\usepackage{amsmath,amssymb,bm,booktabs}
\usepackage[mathscr]{eucal}
\newcommand{\T}{\mathscr{T}}

\newcommand{\C}{\mathscr{C}}
\newcommand{\M}{\mathscr{M}}
\newcommand{\R}{\mathbb{R}}
\renewcommand{\b}{\mathbf}
\usepackage{paralist}
\DeclareMathOperator*{\argmax}{arg\,max}

\begin{document} 

\twocolumn[
\icmltitle{Automated Word Puzzle Generation via Topic Dictionaries}
\icmlauthor{Balázs Pintér}{bli@elte.hu}
\icmlauthor{Gyula Vörös}{vorosgy@inf.elte.hu}
\icmlauthor{Zoltán Szabó}{szzoli@cs.elte.hu}
\icmlauthor{Lőrincz András}{andras.lorincz@elte.hu}
\icmladdress{Faculty of Informatics, E{\"o}tv{\"o}s Lor\'{a}nd University, P{\'a}zm{\'a}ny P. s{\'e}t{\'a}ny 1/C, H-1117 Budapest, Hungary}

\icmlkeywords{natural language processing, word puzzles, topic models, semantic relatedness}

\vskip 0.3in
]

\begin{abstract} 
\end{abstract} 

\section{Introduction}\label{sec:introduction}
Puzzles play a central role in our everyday lives with exciting potentials. As 
assessments in education and psychometry, puzzles are among the most frequently used tools
\cite{verguts2000rasch}. A well-known example is the \emph{odd one out} puzzle of IQ tests 
\cite{carter05iq}. There exist dedicated word puzzles to test or improve a wide array of skills including
language skills, verbal aptitude, logical thinking or general intelligence, such as the 
multiple-choice synonym task of the TOEFL test. Puzzle creation is also a vibrant subfield of
procedural content generation for games (PCG), the automated generation of game content for which 
there is a continuously increasing demand thanks to the thriving popularity of computer and video games.

However, generating and maintaining such puzzles \emph{manually} is quite challenging and expensive:
automated schemes could be of considerable benefit.  
A central problem one has to cope with is variety; otherwise the solver will encounter 
the same (kind of) puzzle multiple times. In case of word puzzles, new puzzles are needed continuously because (i) 
novel words are created on a daily basis (e.g., on blogs), (ii) existing words get new meaning
(e.g., `chat'), (iii) words go out of common use (e.g., `videotape'). Due to the different nature of 
puzzles, the problem of generating puzzles automatically has been tackled only in quite special cases.
There exist, for example efficient techniques for (i) sudoku games, (ii) creating 
mazes on chessboards, or (iii) generating puzzles and quests (objectives for the players) for massively
multiplayer online games. 

To the best of our knowledge, automated \emph{word puzzle} generation is a novel area of this field.
\citet{colton02automated} addressed the problem by a complex theory formation system to
obtain \emph{odd one out}, \emph{analogy} and \emph{next in the sequence} puzzles. The 
presented approach however relied on highly structured datasets, which required serious human annotation effort.

\textbf{Our goal is} to develop a general automated word puzzle generation method from
\begin{compactenum}[1.]
  \item an unstructured and unannotated document collection, i.e., a simple \emph{corpus},
  \item a \emph{topic model}\footnote{Examples include e.g., latent semantic analysis, group-structured
      dictionaries or latent Dirichlet allocation.}, which induces a topic dictionary from the input corpus,
      and
  \item a \emph{semantic similarity} measure of word pairs.
\end{compactenum}

Our method, relying only on these three general components, is capable of (i) generating automatically a large
number of valuable word puzzles of \emph{many different types}, including 
the odd one out, choose the related word and separate the topics puzzle:
\begin{compactitem}[\textbullet]
  \item In \emph{odd one out} puzzles, the solver is required to select the word that is dissimilar to
	the other words.
  \item In \emph{choose the related word} puzzles, the solver has to select the word that is closely
	related to a previously specified group of words.
  \item In \emph{separate the topics} puzzles, the solver has to separate the set of words into two
	disjoint sets of related words.
\end{compactitem}
(ii) The method can create easily \emph{domain-specific} puzzles by replacing the corpus component.
(iii) It is also capable of automatically generating puzzles with \emph{parameterizable levels} suitable for, e.g., beginners or intermediate learners.
In the following, we present the basic ideas behind our approach (Section~\ref{sec:method}) 
with some numerical illustrations (Section~\ref{sec:illustration}). For more extensive demonstrations
and further details, see \cite{pinter12automated2}.

\section{Method}\label{sec:method}
Below, we define the key components of our presented approach for automated word puzzle generation.
The word puzzles we focus on are produced by (i) generating \emph{consistent sets} of related concepts and then
(ii) mixing these sets with weakly related elements: words or other consistent sets. 
For example, in the \emph{odd one out} puzzle, it is sufficient to add a single unrelated word to a consistent set.

For the generation of consistent sets (see Algorithm~\ref{alg:consistent sets}), we assume that we are 
given (i) an \emph{unlabeled corpus} $\b{X}=[\b{x}_1,\ldots,\b{x}_M]\in \R^{N\times M}$ and (ii)
a \emph{topic model} $\T$. In corpus $\b{X}$, the documents ($\b{x}_i\in\R^N$) are represented as weights assigned to words. For example, in a bag of words representation $x_{ij}$ is the number of occurences of the $i$th word in the $j$th document.
Our assumption for the topic model $\T$, is that it induces a \emph{dictionary} 
$\b{D}=\T(\b{X})=[\b{d}_1,\ldots,\b{d}_K]\in\R^{N\times K}$ whose $K$ elements, i.e., \emph{topics} $\b{d}_i$ ($i=1,\ldots,K$) describe well 
the documents in the corpus. 

Numerous topic models ($\T$) fit to this family. 
For example, in latent semantic analysis (LSA; \cite{deerwester90indexing})---which is perhaps the most widely known topic model---the singular value decomposition of $\b{X}=\b{USV}^T$ is computed and
$\b{X}$ is approximated by keeping only the first $K$ (columns of $\b{U}$) left singular vectors; these vectors form $\b{D}$. Group-structured dictionaries
approximate $\b{X}$ by adding a structure-inducing regularization ($\Omega$) on the elements of $\b{D}$, i.e,
minimize the cost function ($\rho\ge 0$)
\begin{align}
  \min_{\b{D}}&\hspace{0.1cm}\frac{1}{\sum_{j=1}^M(j/M)^{\rho}} 
  \sum_{i=1}^M\left(\frac{i}{M}\right)^{\rho}l(\b{x}_i,\b{D}),\\
  l(\b{x},\b{D})&=\min_{\bm{\alpha}}\left[\frac{1}{2}\left\|\b{x}-\b{D}\bm{\alpha}\right\|^2_2+\kappa\Omega(\bm{\alpha})\right]
  \hspace{0.1cm}(\kappa  > 0).
\end{align}
For an excellent review on structured sparsity, see \cite{bach12optimization}. In latent Dirichlet allocation (LDA; \cite{blei03latent})
topics $\b{d}_i\in\R^{N}$ $(i=1,\ldots,K)$ are modelled as latent random variables with a Dirichlet prior. The dictionary 
$\b{D}$ consists of the estimated $\b{d}_i$s.

For word puzzles, we keep only the $k$ ($\le N$) most significant words of the topics as sets:
$m_i=\argmax_k(\b{d}_i)\subseteq\{1,\ldots,N\}$ ($|m_i|=k$). Topic models can produce junk topics \cite{alsumait09topic}. For
example, common function words, such as \emph{did, said}, etc.\ can form a topic. These topics result in
inconsistent sets, whose words are not closely related. To evaluate the consistency of sets and discard
inconsistent ones---which is highly desirable in word puzzles---, we define the consistency of the resulting
word sets $m_i$ using the semantic relatedness of the word pairs they contain measured by explicit semantic analysis
(ESA; \cite{gabrilovich09wikipedia}).  In ESA, given a concept repository, such as the articles of Wikipedia,
the relatedness of two words ($w$, $w'$) is measured as the similarity of their concept based representations:
\begin{equation}
s_{ww'} = \cos(\bm{\varphi}_{ESA}(w),\bm{\varphi}_{ESA}(w')). \label{eq:s}
\end{equation}
The basic assumption of ESA is that if a word appears frequently in a Wikipedia article, then that article
represents the meaning of the word well.

Since in word puzzles, even a single word too weakly connected to the others can make the resulting puzzles ambiguous,
it is prudent to rate each set ($m_i$) according to the \emph{word that is the least related to the other words in the
set}. The semantic relatedness measure (Eq.~\eqref{eq:s}) is also not perfectly accurate: false positives (two
words seem related when in reality they are not) or false negatives (the similarity measure gives a small value
even though the two words are related) may appear.

\begin{figure}[h]
\vskip 0.2in
\begin{center}
  \subfigure[]{\includegraphics[width=3.8cm]{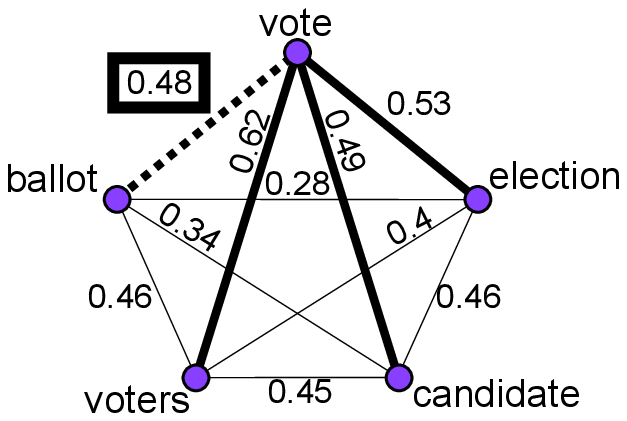}}
  \subfigure[]{\includegraphics[width=4.15cm]{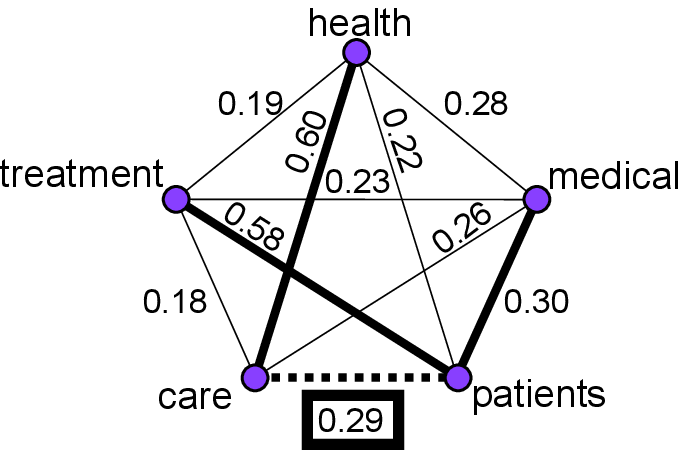}}
  \subfigure[]{\includegraphics[width=3.5cm]{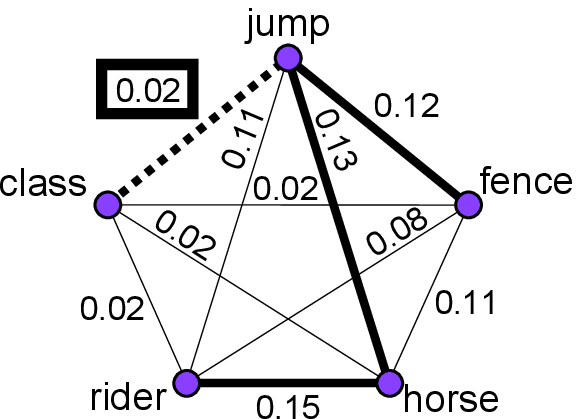}}
  \caption{Checking the consistency of 3 sets of words. Here, sets contain $k=5$ words. Bold edges: maximal spanning tree. 
      Dashed line: edge with minimal weight in the tree; determines consistency. (a): a highly consistent set; all the words are strongly connected to
	the word \emph{vote}.
      (b): a consistent set; some of the relatedness values (e.g., between \emph{care} and \emph{treatment}) are lower than one would expect; The method
	is robust: a relatively high consistency value is assigned to the set. (c): an inconsistent set; the word \emph{class} is weakly connected to the others.\label{fig:set consistency}}
\end{center}
\vskip -0.2in
\end{figure}

To cope with these challenges in determining set ($m$) consistency, one can proceed as follows (see
Algorithm~\ref{alg:consistent sets}, line 7). Robustness to false negatives can be increased by defining the
relatedness of two words based on \emph{all paths} ($\max_{path(i,j)}$) between them in the semantic
similarity ($s_{uv}$, Eq.~\eqref{eq:s}) graph $G=(m,\left.\b{S}\right|_m)$. Robustness to false positives can be
achieved by taking the \emph{minimum relatedness on the path} ($\min_{e\in path(i,j)}s_e$).  Finally, to
ensure that the quality of a set is determined by the two most dissimilar words in the set, one can compute
the minima over all $i\ne j$ word pairs ($\min_{i\ne j}sim(i,j)$). A set is defined to be \emph{consistent} if
this quality of the set is above a given threshold $\delta$.  It can be shown \cite{jungnickel07graphs} that
the $sim(i,j)$ similarity values are equal to the weight of the unique path between $i$ and $j$ in the maximum
spanning tree ($T$) of $G$. Moreover, since the $s_{uv}$ values are non-negative, it is sufficient to find the
edge with minimal weight in $T$ to determine the quality of the set.  For an illustration, see
Fig.~\ref{fig:set consistency}.

\begin{algorithm*}
   \caption{Identify Consistent Sets ($\C$)}
   \label{alg:consistent sets}
    \begin{algorithmic}[1]
      \STATE {\bfseries Input:} corpus $\b{X}=[\b{x}_1,\ldots,\b{x}_M]\in \R^{N\times M}$, 
	  topic model $\T$, size of consistent sets $k$, semantic similarity of words 
	  $\b{S}=[s_{ij}]\in\R^{N\times N}$, consistency treshold $\delta$
      \STATE $\C \gets \emptyset$ \COMMENT{there is no consistent set at the beginning}
      \STATE $\b{D}=\T(\b{X})=[\b{d}_1,\ldots,\b{d}_K]\in\R^{N\times K}$ \COMMENT{compute the topic dictionary}
      \STATE $\M=\{m_1,\ldots,m_K\}$, $m_i=\argmax_k(\b{d}_i)\subseteq\{1,\ldots,N\}$, $|m_i|=k$ \COMMENT{$k$ most significant words of the topics}
      \FORALL{$m \in \M$} 
	  \STATE $G=(m,\left.\b{S}\right|_{m})$ \COMMENT{semantic-weighted graph of the candidate consistent set $m$}
	  \IF[similarity of the 2 most dissimilar words]{$\min\limits_{(i,j)\in m\times m, i\ne j} sim(i,j):=\max\limits_{path(i,j)}\hspace*{0.15cm}\min\limits_{e\in path(i,j)}s_e>\delta$}
	      \STATE $\C \gets \C \cup\{m\}$ \COMMENT{set $m$ is declared to be consistent}
	  \ENDIF
      \ENDFOR
    \end{algorithmic}
\end{algorithm*}

Having the consistent sets ($\C$) at hand, word puzzles can be easily generated by mixing unrelated elements
with $\C$. The pseudocode of \emph{odd one out} puzzle generation is given in Algorithm~\ref{alg:puzzle:odd
one out}. The puzzle generator has two parameters, $\eta_1$ and $\eta_2$. Parameter $\eta_2$ determines
whether the consistent set and the unrelated element are dissimilar enough so that they can be mixed to form a
word puzzle. Parameter $\eta_1$ allows the creation of puzzles of different difficulty (beginner, intermediate,
etc.): by increasing $\eta_1$, the relatedness of the additional elements to the consistent set is increased,
therefore, the puzzle is made harder. Similar constructions can be applied to generate \emph{choose the
related word} or \emph{separate the topics} puzzles \cite{pinter12automated2}.

\begin{algorithm}
   \caption{Odd One Out Puzzle Generation}
   \label{alg:puzzle:odd one out}
    \begin{algorithmic}
      \STATE {\bfseries Input:} consistent sets $\C$, minimal (maximal) relatedness to consistent sets $\eta_1$ ($\eta_2$)
      \FORALL{$C \in \C$} 
	\REPEAT
	\STATE select random word $w$
	\STATE $\sigma\gets\max_{t\in C}\, s_{tw}$ \COMMENT{max.\ relatedness of $w$ to $C$}
	\UNTIL{$\eta_1<\sigma<\eta_2$}
	\STATE output $(C,w)$ puzzle
      \ENDFOR
    \end{algorithmic}
\end{algorithm}

\section{Illustration}\label{sec:illustration}
Here, we illustrate the efficiency of our method in automated \emph{odd one out} puzzle generation.

\begin{figure}
  \begin{center}
  \subfigure[]{\includegraphics[width=4.3cm]{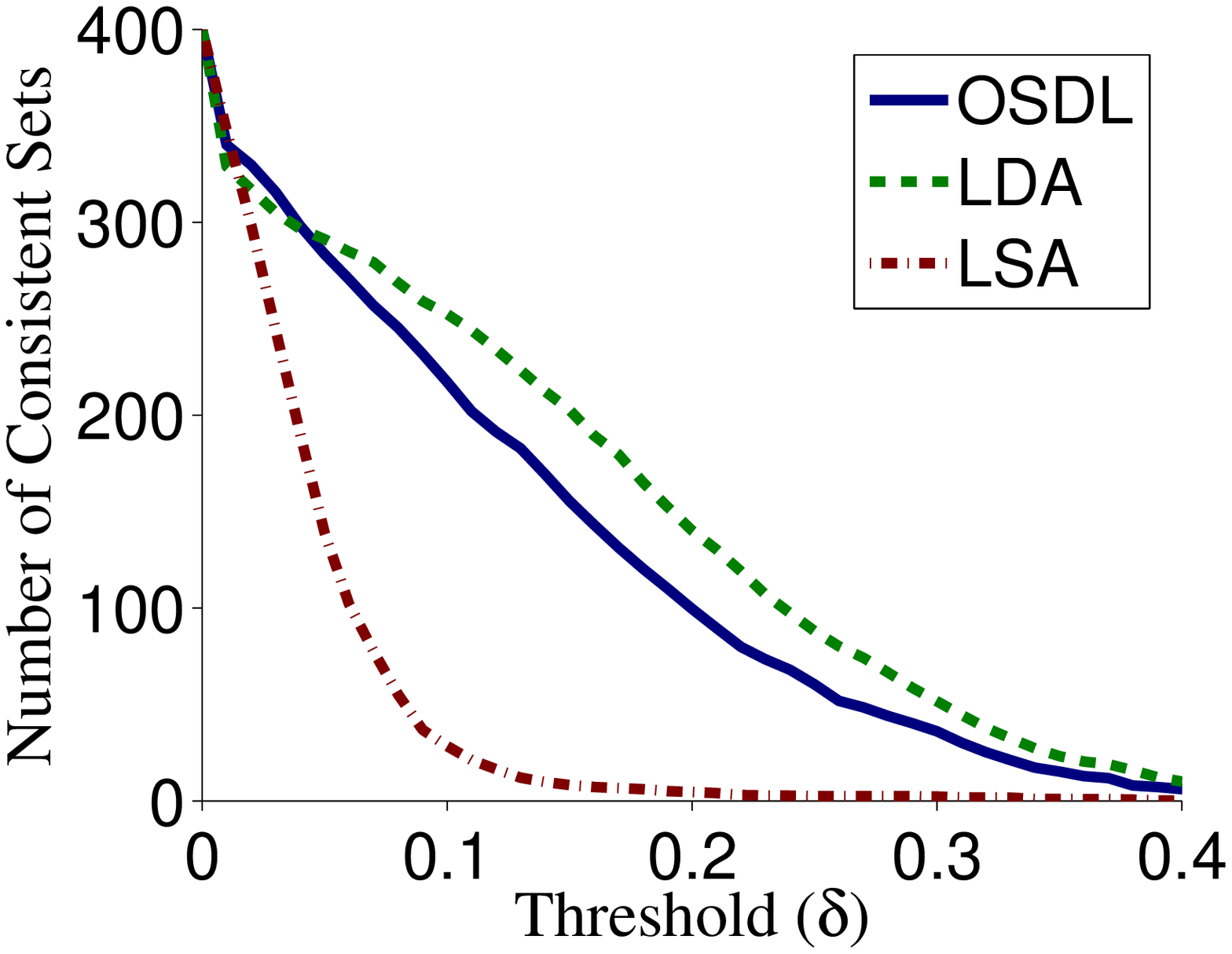}}
  \hspace*{-0.15cm}
  \subfigure[]{\includegraphics[width=4.3cm]{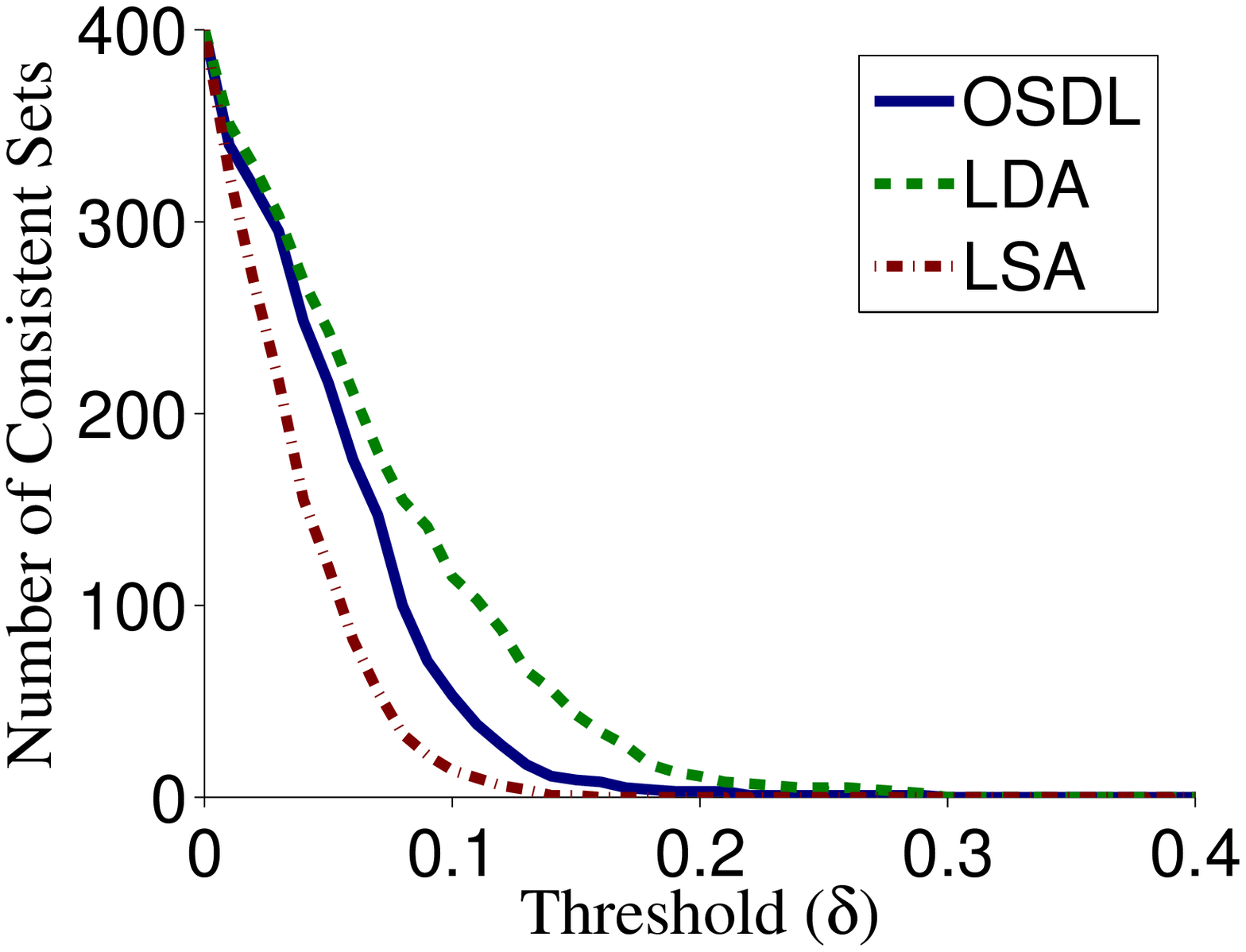}}
  \caption{Number of consistent sets produced by the different topic models as a function of treshold $\delta$. 
  (a): corpus of Wikipedia, (b): NIPS proceedings.\label{fig:numconsistent}}
  \end{center}%
\end{figure}

Consistent sets are a cornerstone of the presented method. 
In the \textbf{first experiment} we 
compared the number of consistent sets of a given quality ($\ge\delta$) 
(i) the different topic models, LSA, LDA and OSDL \cite{szabo11online} a recent group-structured dictionary learning technique could produce,
(ii) on two corpora ($\b{X}$). The two corpora were the English Wikipedia with $M=10,000$ samples and the domain-specific corpus of NIPS proceedings ($M=1,740$).
Consistent sets were composed of $k=4$ words. The number of topics was chosen to be $K=400$. Our results are 
summarized in Fig.~\ref{fig:numconsistent}. According to the figure, out of the studied topic 
models, LDA performs the best, with OSDL following closely behind. LSA does not seem applicable to word 
puzzle generation: it produces very few consistent sets. 
The methods perform better in terms of the number of consistent sets on Wikipedia than on the corpus of NIPS proceedings. 
This is expected, since $M$, the number of articles in the Wikipedia  highly exceeded that of NIPS proceedings.

In the \textbf{second experiment}, we demonstrate the obtained \emph{odd one out} puzzles generated from Wikipedia ($\b{X}$).
We chose $\delta = 0.1$ to obtain a significant number of good enough consistent sets (see Fig.~\ref{fig:numconsistent}). 
First, we illustrate the beginner puzzles (Table~\ref{tab:odd_one_out_beginner}), where we chose parameters 
$\eta_1 = 0.005$, and $\eta_2= 0.02$. In other words, the puzzles generated consist of a consistent set of related words and an unrelated word. 
Beginner puzzles can be solved at first glance by a person who understands the language and has a wide vocabulary, for example, the puzzles \emph{vote, election,
candidate, voters, sony}, or \emph{olympic, tournament, world, championship, acid}. These could be useful for
e.g., beginner language learners or, with a suitable corpus, for children. 
Some puzzles require specific knowledge about a topic. To solve the puzzles \emph{harry, potter, wizard,
ron, manchester} and \emph{superman, clark, luthor, kryptonite, division}, the solver must be familiar with
the book, film, comic, etc. To solve \emph{austria, german, austrian, vienna, scotland}, geographic
knowledge is needed.

\begin{table}[t]
\small
  \begin{center}
  \begin{tabular}{@{}c@{}c@{}c@{}c@{}c@{}}
    \toprule
    \multicolumn{4}{c}{Consistent set of words} & Odd one out \\
    \cmidrule(r){1-4}\cmidrule(l){5-5}
    vote & election & candidate & voters & sony \\
    church & orthodox & presbyterian & evangelical & buddhist \\ 
    olympic & tournament & world & championship & acid \\
    austria & german & austrian & vienna & scotland \\
    devil & demon & hell & soul & boat \\ 
    harry & potter & wizard & ron & manchester \\
    superman & clark & luthor & kryptonite & division \\ 
    magic & world & dark & creatures & microsoft \\
    \bottomrule
  \end{tabular}
  \caption{Odd one out -- beginner puzzles.} \label{tab:odd_one_out_beginner}
  \end{center}
\end{table}

\begin{table}[t]
\small
  \begin{center}
  \begin{tabular}{@{}c@{\hspace{0.15cm}}c@{\hspace{0.15cm}}c@{\hspace{0.15cm}}c@{\hspace{0.15cm}}c@{}}
    \toprule
    \multicolumn{4}{c}{Consistent set of words} & Odd one out \\
    \cmidrule(r){1-4}\cmidrule(l){5-5}
    cao & wei & liu & emperor & king \\
    superman & clark & luthor & kryptonite & batman \\
    devil & demon & hell & soul & body \\
    egypt & egyptian & alexandria & pharaoh & bishop \\
    singh & guru & sikh & saini & delhi \\
    language & dialect & linguistic & spoken & sound \\
    mass & force & motion & velocity & orbit \\
    voice & speech & hearing & sound & view \\
    athens & athenian & pericles & corinth & ancient \\
    function & problems & polynomial & equation & physical \\
    \bottomrule
  \end{tabular}
  \caption{Odd one out -- intermediate puzzles.}\label{tab:odd_one_out_intermediate}
  \end{center}
\end{table}

Second, we illustrate intermediate puzzles (Table~\ref{tab:odd_one_out_intermediate}) 
obtained with $\eta_1 = 0.1$ and $\eta_2 = 0.2$. Although the presented method is based on semantic similarity, it is able to create surprisingly subtle
puzzles. In the puzzle \emph{voice, speech, hearing, sound, view}, the word \emph{view} has a different
modality than the others. To solve the puzzle \emph{cao, wei, liu, emperor, king}, the solver should be
familiar with the three kingdoms period of the Chinese history. For \emph{egypt, egyptian, alexandria, pharaoh,
bishop}, knowledge of the Egyptian history, for \emph{athens, athenian, pericles, corinth, ancient}, familiarity
with the Peloponnesian War is required. In \emph{singh, guru, sikh, saini, delhi}, all the words except
\emph{delhi} are related to sikhism. The puzzle \emph{function, problems, polynomial, equation, physical}
can be solved only with a basic knowledge of mathematics and physics. These results demonstrate the efficiency of our automated word puzzle generation approach.

\textbf{Acknowledgments.}
The European Union and the European Social Fund have provided financial
support to the project under the grant agreement no.\ TÁMOP
4.2.1./B-09/1/KMR-2010-0003. The research has also been supported by the `European Robotic Surgery' 
EC FP7 grant (no.: 288233). Any opinions, findings and conclusions or
recommendations expressed in this material are those of the authors and
do not necessarily reflect the views of other members of the consortium
or the European Commission.

\vspace*{-0.1cm}
\small
\bibliography{main_short}
\bibliographystyle{icml2012}

\end{document}